\documentclass[10pt,twocolumn,letterpaper]{article}

\usepackage{cvpr}
\usepackage{times}
\usepackage{epsfig}
\usepackage{graphicx}
\usepackage{amsmath}
\usepackage{amssymb}

\usepackage{multicol}
\usepackage{multirow}
\usepackage{bm}
\usepackage{authblk}

\usepackage[pagebackref=true,breaklinks=true,letterpaper=true,colorlinks,bookmarks=false]{hyperref}

\cvprfinalcopy 

\def\cvprPaperID{5544} 

\ifcvprfinal\fi
\begin{document}

\title{Look-into-Object: Self-supervised Structure Modeling for Object Recognition}
\renewcommand\Affilfont{\normalsize }
\author[1,2*]{Mohan Zhou}
\author[2*]{Yalong Bai}
\author[2$\dagger$]{Wei Zhang}
\author[1]{Tiejun Zhao}
\author[2]{Tao Mei}
\affil[1]{Harbin Institute of Technology} \affil[2]{JD AI Research, Beijing, China}
\affil[ ]{\tt\small {\{mhzhou99, ylbai\}@outlook.com} {wzhang.cu@gmail.com}  {tjzhao@hit.edu.cn} {tmei@jd.com} }

\newcommand\blfootnote[1]{%
\begingroup 
\renewcommand\thefootnote{}\footnote{#1}%
\addtocounter{footnote}{-1}%
\endgroup 
}

\maketitle

\begin{abstract}
    Most object recognition approaches predominantly focus on learning discriminative visual patterns while overlooking the holistic object structure. Though important, structure modeling usually requires significant manual annotations and therefore is labor-intensive. In this paper, we propose to ``look into object" (explicitly yet intrinsically model the object structure) through incorporating self-supervisions into the traditional framework. We show the recognition backbone can be substantially enhanced for more robust representation learning, without any cost of extra annotation and inference speed.  Specifically, we first propose an object-extent learning module for localizing the object according to the visual patterns shared among the instances in the same category. We then design a spatial context learning module for modeling the internal structures of the object, through predicting the relative positions within the extent. These two modules can be easily plugged into any backbone networks during training and detached at inference time. Extensive experiments show that our look-into-object approach (LIO) achieves large performance gain on a number of benchmarks, including generic object recognition (ImageNet) and fine-grained object recognition tasks (CUB, Cars, Aircraft). We also show that this learning paradigm is highly generalizable to other tasks such as object detection and segmentation (MS COCO). Project page: \url{https://github.com/JDAI-CV/LIO}.
\end{abstract}    
\blfootnote{$*$Equal contribution. This work was done at JD AI research.}
\blfootnote{$\dagger$Corresponding author.}
\section{Introduction} \label{sec:introduction}
Object recognition is one of the most fundamental tasks in computer vision, which has achieved steady progress with the efforts from deep neural network design and abundant data annotations. However, recognizing visually similar objects is still challenging in practical applications, especially when there exist diverse visual appearances, poses, background clutter, and so on.	

\begin{figure}[!t]
    \centering
    \includegraphics[width=\linewidth,page=1]{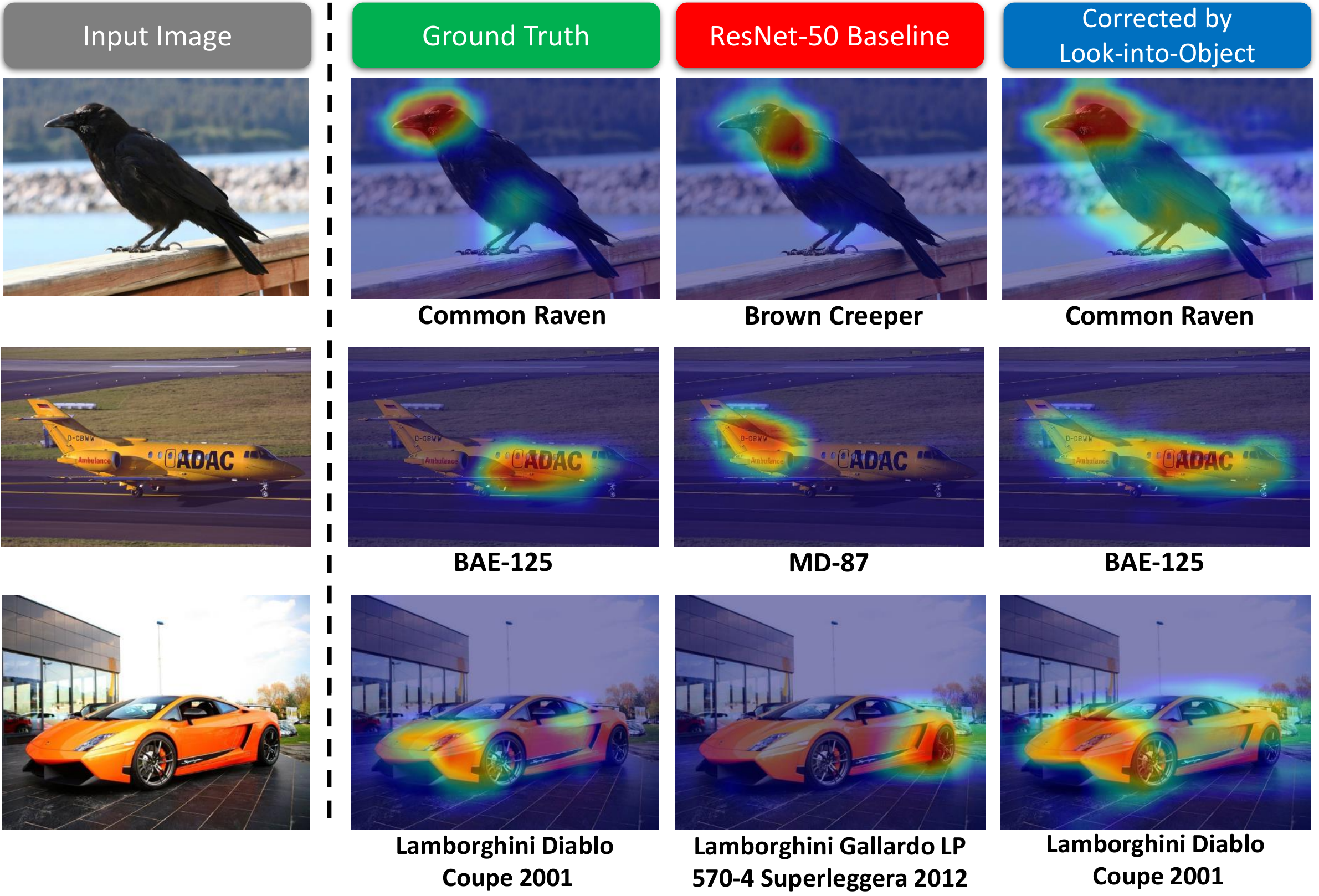}
    \caption{Feature map visualization based on the last convolutional layer of ResNet-50 backbone. The first column shows the original images, while the second and the third columns show the maximally responding feature maps from the ground-truth and the predicted labels, respectively. The last column shows the feature maps by plugging our proposed LIO on ResNet-50. Object extend and discriminative regions are all correctly localized owing to the holistic structure modeling. (Best viewed in color).}
    \label{fig:motivation}
\end{figure}

Suffering from complex visual appearance, it is not always reliable to correctly recognize objects purely based on discriminative regions, even with a large-scale human-labeled dataset. As shown in Fig.~\ref{fig:motivation}, a well-trained ResNet-50 (the third column) can still misclassify objects by looking at the wrong parts.

Existing object recognition approaches can be roughly grouped into two groups. One group optimizes the network architecture to learn high-quality representations~\cite{shih2017deep,BCNN,hu2018squeeze,cui2017kernel}, while the other line of research introduces extra modules to highlight the salient parts explicitly (by bounding-box~\cite{branson2014bird,huang2016part,CoSeq}) or implicitly (by attention~\cite{RACNN,wang2017residual}). Apparently, the latter one costs more on either annotation (\eg bounding boxes / part locations) or calculation (attentions / detection modules). However, all these methods predominantly focus on learning salient patterns while ignoring the holistic structural composition.

In this paper, we argue that correctly identifying discriminative regions largely depends on the holistic structure of objects. Traditional deep learning based methods can be easily fooled in many cases, e.g., distinguishing front and rear tires of a car, localizing legs of a bird among twigs. It is mainly due to the lack of cognitive ability for structures of objects. Therefore, it is crucial to learn the structure of objects beyond simple visual patterns. Though important, it still remains challenging to systematically learn the object structural composition, especially without additional annotation and extra inference time cost.

\begin{figure}[!t]
    \centering
    \includegraphics[width=1\linewidth,page=1]{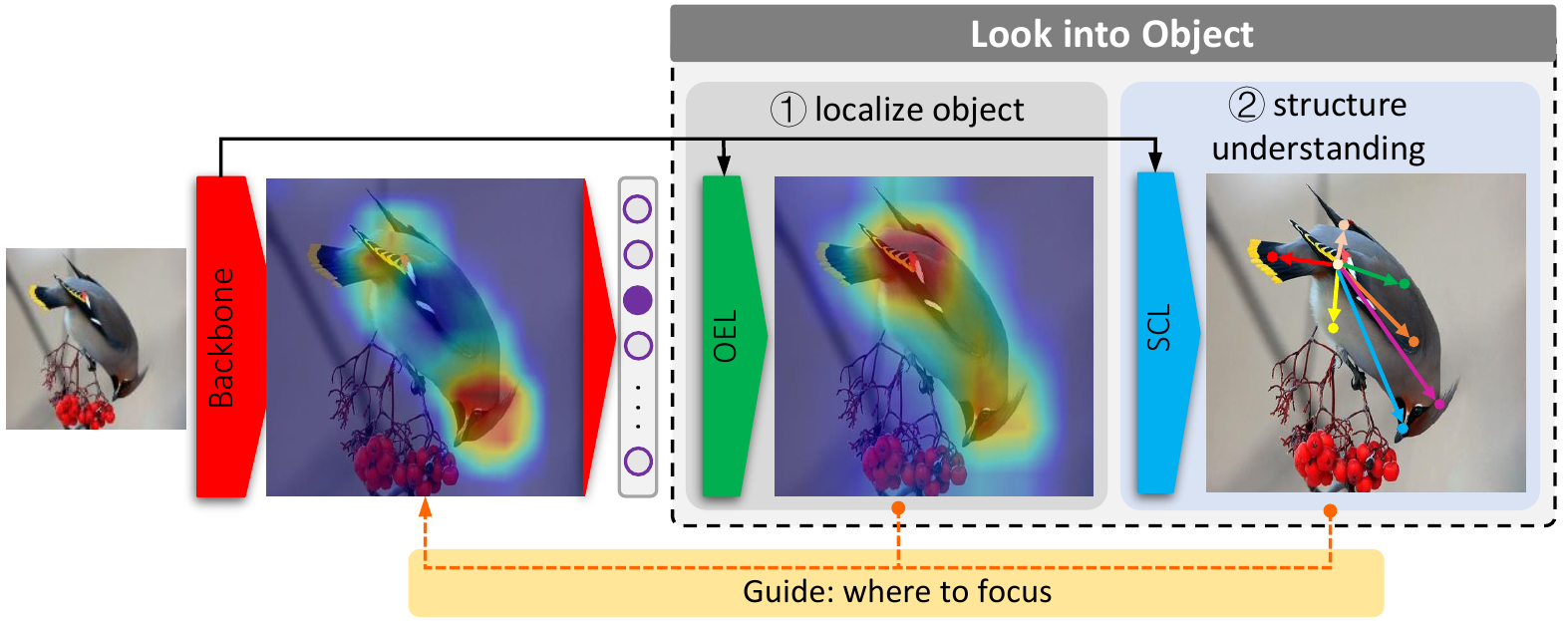}
    \caption{Our proposed Look-into-Object (LIO) approach. Object Extent Learning (OEL) and Spatial Context Learning (SCL) enforce the backbone to learn object extent and internal structure respectively.}
    \label{fig:intro}
\end{figure}

In this work, we propose to model the holistic object structure without additional annotation and extra inference time. Specifically, we propose to ``look-into-objects" (in short ``LIO’’) to understand the object structure in images by automatically modeling the context information among regions. From the psychological point of view, recognizing an object can be naturally regarded into two stages: 1) roughly localizing the object extent (the whole extent of the object rather than object part) in the image, and 2) parsing the structure among parts within the object.

Accordingly, we design two modules to mimic such a psychological process of object recognition. We propose a novel and generic scheme for object recognition by embedding two additional modules into a traditional backbone network, as shown in Fig.~\ref{fig:intro}. The first one is \textit{Object-Extent Learning Module} (OEL) for object extent localization, while the second is \textit{Spatial Context Learning Module} (SCL) for structure learning within the object.

Naturally, a prerequisite for object structure modeling is that the object extent can be localized. The OEL module enforces the backbone to learn object extent using a pseudo mask. We first measure the region-level correlation between the target image and other positive images in the same category. The regions belonging to the main object would have high correlations, owing to the commonality among images from the same category. As a result, a pseudo mask of object extent can be constructed according to the correlation scores without additional annotation besides the original image labels. Then, the backbone network is trained to regress the pseudo mask for localizing the object. With the end-to-end training, the capacity of object-extent localization for backbone network can be further reinforced.

The SCL module predicts the spatial relationships among regions within the object extent in a self-supervised manner. Given the localized extent learned by the OEL module, the SCL mainly focuses on the internal structure among regions. Specifically, we enforce the backbone network to predict the relative polar coordinates among pairs of regions, as shown in Fig.~\ref{fig:intro}. In this way, the structural composition of object parts can be modeled. This self-supervised signal can benefit the classification network for the object structure understanding by end-to-end training. Obviously, localize the discriminative regions in a well-parsed structure is much easier than in the raw feature maps.

Note that all these modules take the feature representations generated by the classification backbone network as input and operate at a regional level, which leads to a delicate \textit{Look-into-Object} (LIO) framework. Training with such objectives enforces the feature learning of the backbone network by the end-to-end back-propagation. Ideally, both object extent and structure information can be injected into the backbone network to improve object recognition without additional annotations. Furthermore, both modules can be disabled during inference time.

The main contributions can be summarized as follows:

1. A generic LIO paradigm with two novel modules: object-extent learning for object-extent localization, and self-supervised spatial context learning module for modeling object structural compositions.

2. Experimental results on generic object recognition, fine-grained recognition, object detection, and semantic segmentation tasks demonstrate the effectiveness and generalization ability of LIO.

3. From the perspective of practical application, our proposed methods do not need additional annotation and introduce no computational overhead at inference time. Moreover, the proposed modules can be plugged into any CNN based recognition models.
\section{Related Work} \label{sec:related}

\noindent{\textbf{Generic Object Recognition:}}
General image classification was popularized by the appearance of ILSVRC~\cite{ILSVRC}. With the extraordinary improvement achieved by AlexNet~\cite{AlexNet}, deep learning wave started in the field of computer vision. Since then, a series works, e.g. VGGNet~\cite{VGGNet}, GoogLeNet~\cite{inception}, ResNet~\cite{resnet}, Inception Net~\cite{inception, inceptionv2}, SENet~\cite{hu2018squeeze}, etc. are proposed to learn better representation for  image recognition.

However, general object recognition models still suffer from easy confusion among visually similar objects~\cite{bilal2018convolutional, deng2010what}. The class confusion patterns usually follow a hierarchical structure over the classes. General object recognition networks usually can well separate high-level groups of classes, but it is quite costly to learn specialized feature detectors that separate individual classes. The reason is that the global geometry and appearances of the classes in the same hierarchy can be very similar. As a result, how to identify their subtle differences in the discriminative regions is of vital importance.

\noindent{\textbf{Fine-Grained Object Recognition:}}
Different from general object recognition, delicate feature representation of object parts play a more critical role in fine-grained object recognition. Existing fine-grained image classification methods can be concluded in two directions. The first one is to enhance the detailed feature representation ability of the backbone network~\cite{F1, F2, F3}. The second one is to introduce part locations or object bounding box annotations as an additional optimization objective or supervision besides basic classification network~\cite{DVAN, MACNN, RACNN, CoSeq}.

Similar to general object recognition, deep learning based feature representations achieved great success on fine-grained image recognition~\cite{donahue2014decaf,sharif2014cnn}. After that, second-order bilinear feature representation learning methods~\cite{BCNN} and a series of extensions~\cite{wei2018grassmann,kong2017low,yu2018hierarchical} were proposed for learning local pairwise feature interactions in a translation invariant manner. 

However, recognizing objects from a fine-grained category requires the neural network to focus more on the discriminative parts~\cite{PARTPROVE}. To address this problem, a large amount of part localization based fine-grained recognition methods are proposed. Most of these methods applied attention mechanism to obtain discriminative regions~\cite{RACNN, MAMC}. Zheng~\etal~\cite{MACNN} tried to generate multiple parts by clustering, then classified these parts to predict the category. Compared with earlier part based methods, some recent works tend to use weak supervisions or even no annotation of parts or key areas~\cite{OPAM,yang2018learning}. In particular, Peng~\etal~\cite{OPAM} proposed a part spatial constraint to make sure that the model could select discriminative regions, and a specialized clustering algorithm is used to integrate the features of these regions. Yang~\etal~\cite{yang2018learning} introduced a method to detect informative regions and then scrutinizes them for final predictions. These previous works aim to search for key regions from pixel-level images directly. However, to correctly detect discriminative parts, the deep understanding of the structures of objects and the spatial contextual information of key regions are essential. In turn, the location information of regions in images can enhance the visual representation of neural networks~\cite{PUZZ}, which has been demonstrated on unsupervised feature learning.

Different from previous works, our proposed method focuses on modeling spatial connections among object parts for understanding object structure and localizing discriminative regions. Inspired by the studies that contextual information among objects influences the accuracy and efficiency of object recognition~\cite{hock1974contextual}, the spatial information among regions within objects also benefits the localization of discriminative regions. Thus we introduce two modules in our proposed method; the first one aims to detect the main objects, and the second one inferences the spatial dependency among regions in objects. The experimental results show that our method can improve the performance of both general object recognition and fine-grained object recognition. Moreover, our method has no additional overhead except the backbone network feedforward during inference.

\section{Approach} \label{sec:method}
\begin{figure*}[!ht]
    \centering
    \includegraphics[width=\textwidth,page=1]{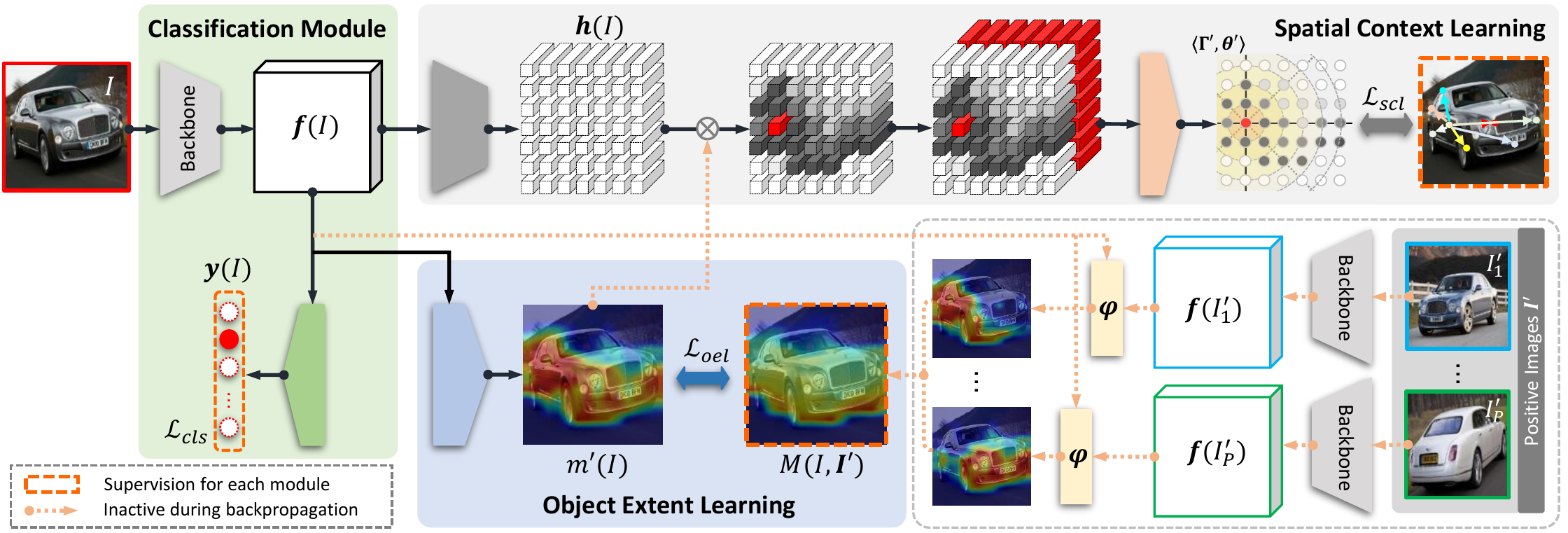}
    \caption{The overall pipeline of our Look-into-object (LIO) framework. The feature maps $\bm{f}(I)$ extracted from the classification module are further fed into spatial context learning module and object-extent learning module. After end-to-end training, the backpropagation signals from spatial context learning module and object-extent learning module can jointly optimize the representation learning of the backbone network in classification module. Only the classification module (in the green box) is activated during inference.} 
    \label{fig:framework}
\end{figure*}

In this section, we introduce our proposed LIO approach. As shown in Fig.~\ref{fig:framework}, our network is mainly organized by three modules: 
\begin{itemize}\setlength\itemsep{-0.2em}
    \item \textbf{Classification Module} (CM): the backbone classification network that extracts basic image representations and produces the final object category.
    \item \textbf{Object-Extent Learning Module} (OEL): a module for localizing the main object in a given image.
    \item \textbf{Spatial Context Learning Module} (SCL): a self-supervised module to strengthen the connections among regions through interactions among feature cells in CM. 
\end{itemize}

Given an image $I$ and its ground truth one-hot label $\bm{l}$, we can get the feature maps $\bm{f}(I)$ of size $N\times N\times C$ from one of the convolutional layers, and the probability vector $\bm{y}(I)$ from the classification network. $C$ is the channel size of that layer, and $N\times N$ is the size of each feature map in $\bm{f}(I)$. The loss function of the classification module (CM) $\mathcal{L}_{cls}$ can be written as:
\begin{equation}
    \mathcal{L}_{cls} = -\sum_{I\in \mathcal{I}}\bm{l}\cdot \log \bm{y}(I),
\end{equation}
where $\mathcal{I}$ is the image set for training.

The object-extent learning module and spatial context learning module are designed to help our backbone classification network learn representations beneficial to structure understanding and object localization. These two modules are light-weighted, and only a few learnable parameters are introduced. Furthermore, OEL and SCL are disabled at inference time, and only the classification module is needed for computational efficiency.

\subsection{Object-Extent Learning (OEL)}

Localizing the extent of the object in an image is a prerequisite for understanding the object structure. A typical approach is to introduce bounding boxes or segmentation annotations, which cost much on data collection. For typical image recognition task that lacks localization or segmentation annotations, 
we propose a new module called \textit{Object-Extent Learning} to help the backbone network distinguish the foreground and background.

We can partition the feature maps $\bm{f}(I)$ into $N \times N$ feature vector $\bm{f}(I)_{i,j}\in\mathbb{R}^{1\times C}$, where $i$ and $j$ are the horizontal and vertical indices respectively ($1 \leq i, j \leq N$). Each feature vector centrally responds to a certain region in input image $I$. 

Inspired by the principle that objects in the image from the same category always share some commonality, and the commonality, in turn, help the model recognize objects, we sample a positive image set $\bm{I}' = \{I_1', I_2', \cdots, I_P'\}$ with the same label $\bm{l}$ of image $I$, and then measure the region-level correlations between $\bm{f}(I)_{i,j}$ and each image $I'\in\bm{I}'$ by
\begin{equation}
    \label{equ:correlation}
    \varphi_{i,j}(I, I')
    = \frac{1}{C}\max_{\substack{1 \leq i',j' \leq N}}  {\langle \bm{f}(I)_{i, j}, {\bm{f}(I')_{i', j'}} \rangle},
\end{equation}
where $\langle \cdot,\cdot \rangle$ denotes dot product.

Jointly trained with the classification objective $\mathcal{L}_{cls}$, the correlation score $\varphi_{i,j}$ is usually positively correlated with the semantic relevance to $\bm{l}$. 

After that, we can construct a $N \times N$ semantic mask matrix $\varphi(I,I')$ for the object extent in $I$.

\begin{figure}[!ht]
    \centering
    \includegraphics[width=\linewidth,page=1]{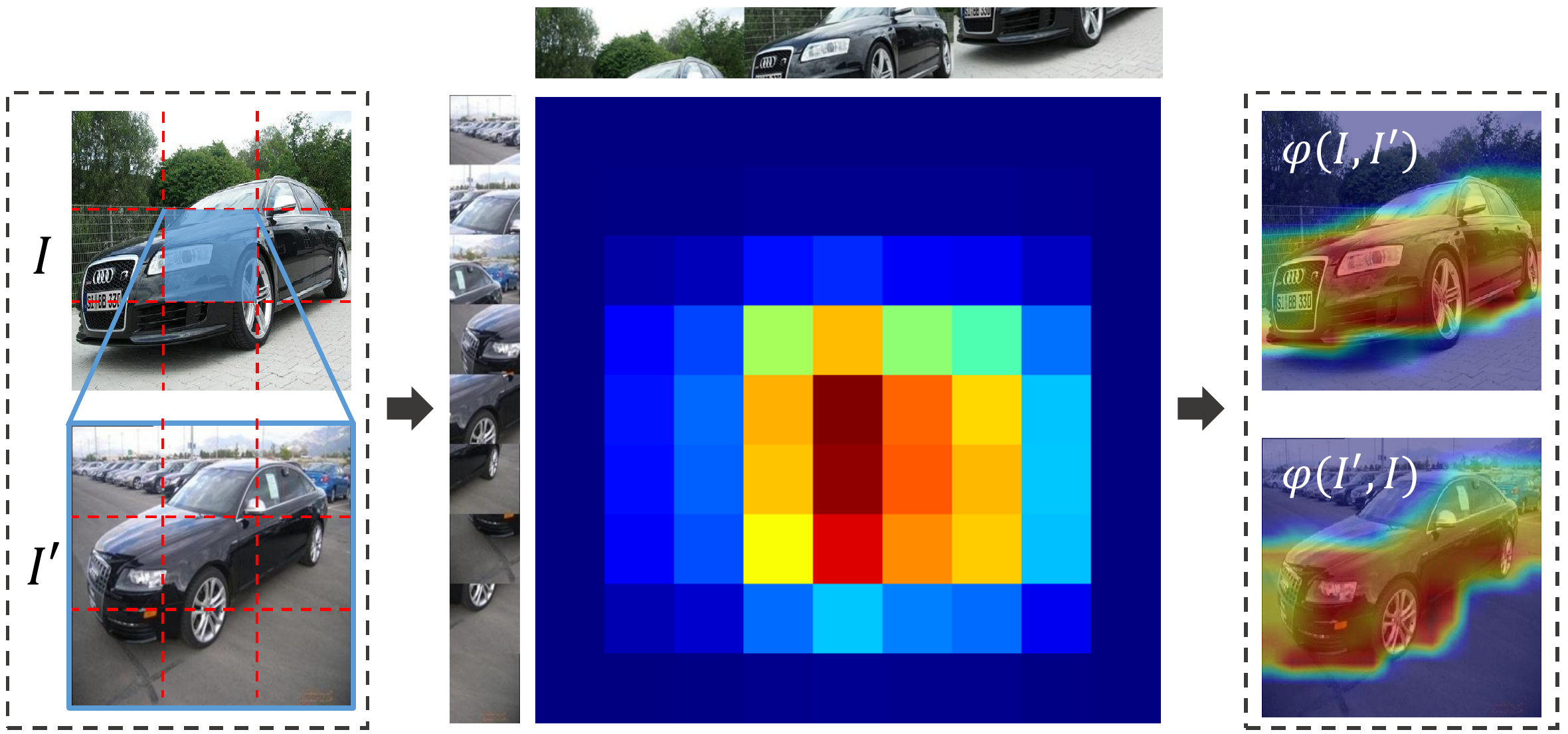}
    \caption{Correlation calculation helps to localize object extent.} 
    \label{fig:wcm_illustrate}
\end{figure}

Therefore, the commonality of images from the same category can be well captured by this semantic correlation mask $\varphi$, and the values in $\varphi$ distinguish the main object area and background naturally, as shown in Fig.~\ref{fig:wcm_illustrate}. 

Taking the impact of viewpoint variation and deformation into account, we use multiple positive images to localize the main area of an object. Therefore, we get a weakly supervisory pseudo label to mimic the object localization masks:
\begin{equation}
    M(I, \bm{I}') = \frac{1}{P}\sum_{p=1}^P \varphi(I, I'_p).
\end{equation}
Also, $M(I, \bm{I}')$ can be regarded as the representations of the commonality shared among images from the same category. 

The primary purpose of the OEL module is to enrich the classification network from the commonality and infer the semantic mask of the object extent. Thus we equip a simple stream after $\bm{f}(I)$ to fuse all feature maps in $\bm{f}(I)$ with weights. The features are processed by a $1\times 1$ convolution to obtain outputs with one channel $m'(I)$. Different from traditional attention that aims to detect some specific parts or regions, our OEL module is trained for gathering all regions within the object and neglect the background or other irrelevant objects. 

The loss of OEL module $\mathcal{L}_{oel}$ can be defined as the distance between pseudo mask $M(I,\bm{I}')$ of the object extent and $m'(I)$, which can be expressed as:

\begin{equation}
    \mathcal{L}_{oel} 
    = \sum_{I\in\mathcal{I}}\mbox{MSE}\big(m'(I), M(I, \bm{I}')\big),
\end{equation}
where $\mbox{MSE}$ is defined as a mean-square-error loss function.

$\mathcal{L}_{oel}$ is helpful to learn a better representation of the object extent according to visual commonality among images in the same category.
By end-to-end training, the object-extent learning module can enrich the backbone network by detecting the main object extent.

\subsection{Spatial Context Learning (SCL)}
Structural information plays a significant role in image comprehension. Classical general convolutional neural networks use convolutional kernels to extract structural information in the image, and fuse the multi-level information by stacking layers. We propose a self-supervised module called \textit{Spatial Context Learning} to strengthen the structural information for the backbone network by learning the spatial context information in objects. 

Given an image $I$, our SCL module also acts on the feature maps $\bm{f}(I)$ and aims to learn the structural relationships among regions. Firstly, the feature map is processed by a $1\times 1$ convolution plus a ReLU such that we get the new map $\bm{h}(I)\in\mathbb{R}^{N\times N\times C_1}$, describing the spatial information of different feature cells. Each cell in $\bm{h}(I)$ centrally represents the semantic information of an area of the image $I$. The structural relationships among different parts of an object can be easily modeled by building spatial connections among different regions.

In this paper, we apply polar coordinates for measuring the spatial connections among different regions. Given a reference region $R_o = R_{x,y}$ whose indices are $\left(x,y\right)$ in $N\times N$ plane, and a reference horizontal direction, the polar coordinates of region $R_{i,j}$ can be written as $(\Gamma_{i,j}, \theta_{i,j})$:
\begin{equation}
    \begin{aligned}
    \Gamma_{i,j} &= \sqrt{(x-i)^2+(y-j)^2}/\sqrt{2}N\\
    \theta_{i,j} &= (\bm{\mathrm{atan2}}(y-j,x-i) + \pi) / 2\pi,
    \end{aligned}
\end{equation}
where $0 < \Gamma_{i,j} \leq 1$ measures the relative distance between $R_o$ and $R_{i,j}$, $\bm{\mathrm{atan2}(\cdot)}$ returns a unambiguous value in range of $(-\pi,\pi]$ for the angle converting from Cartesian coordinates to polar coordinates, and $\theta_{i,j}$ measures the polar angle of $R_{i,j}$ corresponding to the horizontal direction. It is worth noting that, to ensure a wide range of the distribution of the values of $\theta$, ideally, the region within the object extent should be selected as the reference region. In this paper, the region who respond to the maximum value in $m(I)$ is selected:
\begin{equation}
    R_o = R_{x,y}, \text{where}\left(x,y\right)=\arg\max_{1\leq x, y\leq N}m'(I)_{i,j}
\end{equation}

This ground-truth polar coordinates is regarded as supervision for guiding the SCL module training. Specifically, the SCL module is designed for predicting the polar coordinates of region $R_{i,j}$ by jointly considering the representations of target region $R_{i,j}$ and reference region $R_{o}$ from $\bm{h}(I)$. 
We first apply channel-wise concatenation for $h(I)_{i,j}$ and $h(I)_{x,y}$, then the outputs are handled by a fully-connected layer with ReLU to get the predicted polar coordinates $(\Gamma_{i,j}', \theta_{i,j})'$. Since our proposed modules mainly focus on modeling the spatial structures of different parts within the object, the object-extent mask $m'(I)$ learned from the OEL module is also adapted in the SCL module.

There are two objectives in the SCL module. The first one measures the relative distance differences of all regions with object:
\begin{equation}
    \mathcal{L}_{dis} = \sum_{I\in\mathcal{I}}\sqrt{\frac{\sum_{1\leq i,j\leq N}m'(I)_{i,j}(\Gamma_{i,j}' - \Gamma_{i,j})^2}{\sum m'(I)}}.
\end{equation}
The other one measures the polar angle differences of regions inside the object. Considering the structural information for an object should be rotation invariant, and robust to various appearances and poses of the object, we measure the polar angle difference $\mathcal{L}_{\angle}$ according to the \textit{standard deviation} of gaps between predicted polar angles and ground-truth polar angles:
\begin{equation}
\begin{aligned}
    \mathcal{L}_{\angle} &= \sum_{I\in\mathcal{I}}\sqrt{\frac{\sum_{1\leq i,j\leq N}m'(I)_{i,j}\left( \theta_{\Delta_{i,j}} - \overline{\theta}_{\Delta} \right )^2}{\sum m'(I)}},\\
    \theta_{\Delta_{i,j}} &= \begin{cases}
        \theta'_{i,j} - \theta_{i,j}, & \text{if } \theta'_{i,j} - \theta_{i,j} \geq 0\\
         1+\theta'_{i,j} - \theta_{i,j}, & 
        \text{otherwise},
    \end{cases}\\
\end{aligned}
\end{equation}
where $\overline{\theta}_{\Delta} = \frac{1}{\sum m'(I)}\sum_{1\leq i,j\leq N}{m'(I)_{i,j}\theta_{\Delta_{i,j}} }$ is the mean of the gaps between predicted polar angles and ground-truth polar angles. In this way, our SCL could focus on modeling the relative structure among parts of the object rather than the absolute position of regions that is sensitive to the reference direction. Moreover, owing to the usage of predicted semantic mask $m'(I)$, other visual information except for the main object, e.g., background, is ignored during regressing polar coordinates. 

Overall, the loss function of the \textit{Spatial Context Learning Module} can be written as:
\begin{equation}
    \mathcal{L}_{scl} = \mathcal{L}_{dis} + \mathcal{L}_{\angle}.
\end{equation}
With $\mathcal{L}_{scl}$, the backbone network can recognize the pattern structures, i.e., the composition of the object. By end-to-end training, the spatial context learning module can empower the backbone network to model the spatial dependence among parts of the object.

\subsection{Joint Structure Learning}
In our framework, the classification, object-extent learning and spatial context learning modules are trained in an end-to-end manner, in which the network can leverage both enhanced object localization and object structural information. The whole framework is trained by minimizing the following objective:
\begin{equation}
    \label{equ:loss_f}
    \mathcal{L} = \mathcal{L}_{cls} + \alpha\mathcal{L}_{oel} + \beta\mathcal{L}_{scl}.
\end{equation}
We set $\alpha=\beta=0.1$ for all experimental results reported in this paper.

During inference, both the SCL and OEL are removed, and only the Classification Module is kept. Thus, the framework does not introduce additional computational overhead at inference time and runs faster for practical product deployment.

Moreover, the object-extent learning module and spatial context learning module can be attached to different stages of feature maps generated from different convolutional layers of the Classification Module. Thus we can model the structural information of the object in different granularity levels. Together, the overall training method is named as \textbf{multi-stage LIO}. For example, we can jointly optimize our framework by the combination of $\mathcal{L}_{7\times 7}$ (extracting feature maps from the last convolutional layer with $N=7$) and $\mathcal{L}_{14\times 14}$ (from the penultimate convolutional layer with $N=14$) for ResNet-50.

\section{Experiments} \label{sec:experiments}
To show the superiority of our proposed look-into-object framework, we evaluate the performance on two object recognition settings: fine-grained object recognition and generic image classification. Furthermore, we also explore our LIO framework in other tasks, such as object detection and segmentation, to study its generalization ability. 

Unless specially mentioned, the spatial context learning module and object-extent learning module are applied on the feature map of the last stage in the backbone classification network, and three positive images are used for training procedure by default. 
For all of these tasks, we did not use any additional annotations.

\subsection{Fine-grained Object Recognition}
For fine-grained object recognition, we test LIO on three different standard benchmarks: CUB-200-2011 (CUB)~\cite{CUB}, Stanford Cars (CAR)~\cite{CAR} and FGVC-Aircraft (AIR)~\cite{AIR}.

\begin{table}[!t]
\small
\begin{center}
    \begin{tabular}{|l|c|c|c|}
    \hline
    \multicolumn{1}{|c|}{\multirow{2}{*}{Method}}  & \multicolumn{3}{c|}{Accuracy (\%)}  \\ \cline{2-4} 
    \multicolumn{1}{|c|}{} & CUB & CAR & AIR \\ \hline\hline
    CoSeq (+BBox)~\cite{CoSeq} & 82.8         & 92.8          & -             \\
    FCAN (+BBox)~\cite{liu2016fully}   & 84.7         & 93.1          & -             \\\hline
    B-CNN~\cite{BCNN}  & 84.1         & 91.3          & 84.1          \\
    HIHCA~\cite{cai2017higher}  & 85.3         & 91.7          & 88.3          \\
    RA-CNN~\cite{RACNN}  & 85.3         & 92.5          & 88.2          \\
    OPAM~\cite{OPAM}  & 85.8         & 92.2          & -             \\
    Kernel-Pooling~\cite{cui2017kernel}  & 84.7         & 91.1          & 85.7          \\
    MA-CNN~\cite{zheng2017learning}  & 86.5         & 92.8          & 89.9          \\
    DeepKSPD-rootm~\cite{engin2018deepkspd} & 86.5         & 93.2          & 91.0          \\
    MAMC~\cite{MAMC}   & 86.5         & 93.0          & -             \\
    HBP~\cite{yu2018hierarchical}  & 87.1         & 93.7          & 90.3          \\
    DFL-CNN~\cite{wang2018learning} & 87.4         & 93.1          & 91.7          \\
    NTS-Net~\cite{yang2018learning}  & 87.5         & 93.9          & 91.4          \\
    DCL~\cite{chen2019destruction} & 87.8 & \textbf{94.5} & \textbf{93.0} \\
    \hline\hline
    ResNet-50 Baseline  & 85.5 & 92.7 & 90.3 \\
    \hline
    LIO/ResNet-50 ($7\times 7$)    & 87.3         & 93.9           & 92.4          \\ 
    LIO/ResNet-50 ($14\times 14$)    & 87.3         & 94.2            & 92.3          \\
    LIO/ResNet-50 ($28\times 28$)    & 87.6         & 94.0         & 92.4          \\
    LIO/ResNet-50 (multi-stage)    & \textbf{88.0}&    \textbf{94.5}          &    92.7          \\
    \hline
    \end{tabular}
\end{center}
\caption{Comparison results on three different fine-grained object recognition benchmarks.}
\label{tab:exp-fine-grained}
\end{table}

\begin{figure*}[!ht]
    \centering
    \includegraphics[width=\linewidth,page=1]{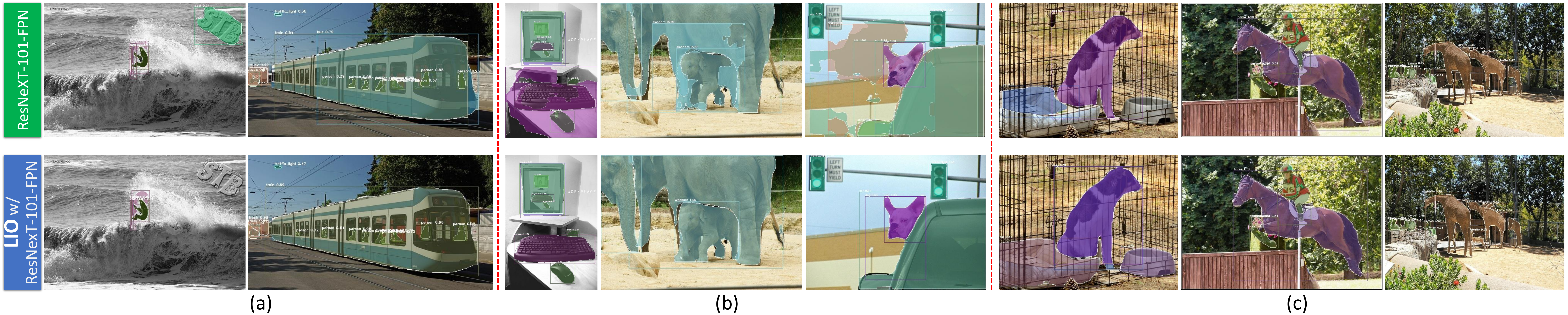}
    \caption{Qualitative examples for COCO object detection and instance segmentation. Our LIO based method can help improve the performance according to object structure information in three aspects: (a) reducing incorrect object label prediction. (b) neglecting noisy segmentation mask. (c) completing fragmentary segmentation mask. Best viewed in electronic version.}
    \label{fig:maskrcnn_visualization}
\end{figure*}

We first initialize LIO with ResNet-50 backbone pre-trained on ImageNet classification task, and then finetune our framework on the datasets above-mentioned. The input images are resized to a fixed size of $512\times512$ and randomly cropped into $448\times448$ for scale normalization. We adopt random rotation and horizontal flip for data augmentation. All above transformations are standard in the literature. Both ResNet-50 baseline and LIO/ResNet-50 are trained for 240 epochs to ensure complete convergence. SGD is used to optimize the training loss as defined in Equation~\ref{equ:loss_f}. 
During testing, only the backbone network is applied for classification. The input images are centrally cropped and then fed into the backbone classification network for final predictions.

Detailed results are summarized in Table~\ref{tab:exp-fine-grained}. Besides plugging the OEL and SCL to the last stage feature map of size $7\times 7$, we also tested these two modules on the penultimate stage $14\times 14$ output, and the antepenultimate stage $28\times 28$ output. Then these three different stages of models are combined into a multi-stage LIO. As in Table~\ref{tab:exp-fine-grained}, the LIO embedded ResNet-50 can achieve significantly better accuracy than baseline ResNet-50. Moreover, the multi-stage LIO achieves significant performance improvements on all three benchmarks, which proves the effectiveness of the proposed region-level structure learning framework. 

It worthy note that LIO and our previous work DCL~\cite{chen2019destruction} target at different research lines in the fine-grained recognition task. DCL aims to learn \textit{discriminative local regions}, while LIO tries to understand the \textit{structure of the whole object}. Both of these two kinds of methods can benefit the fine-grained object recognition, while LIO works better on recognition of flexible objects (CUB), and can be further expanded into generic object recognition (Sec.~\ref{sec:generic_recog}), object detection and segmentation (Sec.~\ref{sec:det_seg}) since object structure information plays an essential role in those tasks.

\subsection{Generic Object Recognition on ImageNet}\label{sec:generic_recog}
We also evaluate the performance of our proposed LIO on large-scale general object recognition dataset ImageNet-1K (ILSVRC-2012), which including 1.28 million images in 1000 classes. For compatibility test, we evaluate our method on the commonly used backbone network ResNet-50. Following standard practices, we perform data augmentation with random cropping to a size of $224\times224$ pixels and perform random horizontal flipping. 
The optimization is performed using SGD with momentum 0.9 and a minibatch size of 256. 

The experimental results are reported in Table~\ref{tab:exp-imagenet}. We can find that LIO boosts the performance of three different backbone networks on the ImageNet-1K validation set, which further demonstrates the generality ability of our proposed object recognition framework. With a lightweight LIO plugin, the performance of typical ResNet-50 can even achieve the performance of SE-ResNet-50~\cite{hu2018squeeze}.

\begin{table}[!t]
\small
\begin{center}
    \begin{tabular}{|l|cc|}
    \hline
    Method          & Top-1 err. & Top-5 err. \\
    \hline\hline
    ResNet-50~\cite{resnet}       & 24.80 & 7.48      \\
    LIO/ResNet-50 ($7\times 7$)   & 23.63 & 7.12      \\
    LIO/ResNet-50 ($14\times 14$) & 23.60 & 7.10      \\
    LIO/ResNet-50 (multi-stage)   & \textbf{22.87} & \textbf{6.64}\\
    \hline
    \end{tabular}
\end{center}
\caption{Single-crop error rates (\%) of single model on the ImageNet-1K validation set.}
\label{tab:exp-imagenet}
\end{table}

\begin{table*}[!h]
\small
\begin{center}
    \begin{tabular}{|l|cccccc|cccccc|}
    \hline
    \multirow{2}{*}{Method} & \multicolumn{6}{c|}{Object Detection} & \multicolumn{6}{c|}{Semantic Segmentation} \\ \cline{2-13}
    & $AP$ & $AP_{50}$ & $AP_{75}$ & $AP_S$ & $AP_M$ & $AP_L$ 
    & $AP$ & $AP_{50}$ & $AP_{75}$ & $AP_S$ & $AP_M$ & $AP_L$ \\
    \hline\hline
    ResNet-50-C4 & 35.9 & 56.1 & 38.9 & 18.0 & 40.1 & 49.7
                 & 31.5 & 52.8 & 33.0 & 12.1 & 34.7 & 49.3 \\
    LIO/ResNet-50-C4  & \textbf{37.6} & \textbf{57.5} & \textbf{41.0} 
                      & \textbf{21.0} & \textbf{41.8} & \textbf{52.0}
                      & \textbf{32.6} & \textbf{54.1} & \textbf{34.7} 
                      & \textbf{14.3} & \textbf{35.7} & \textbf{51.3} \\
    \hline\hline
    ResNeXT-101-FPN & 41.1 & 62.8 & 45.0 & 24.0 & 45.4 & 52.6 
                    & 37.1 & 59.4 & 39.7 & 17.7 & 40.5 & 53.8 \\
    LIO/ResNeXT-101-FPN & \textbf{42.0} & \textbf{63.3} & \textbf{46.0} 
                           & \textbf{24.7} & \textbf{46.1} & \textbf{54.3} 
                           & \textbf{37.9} & \textbf{60.0} & \textbf{40.6} 
                           & \textbf{18.1} & \textbf{41.1} & \textbf{54.8} \\
    \hline
\end{tabular}

\end{center}
\caption{Object detection and segmentation results on COCO {\tt val2017} set.}
\label{tab:exp-detection}
\end{table*}

\subsection{Object Detection and Segmentation on COCO}\label{sec:det_seg}

Meanwhile, considering the object structure information would be helpful for object detection and segmentation tasks, we also investigate our proposed LIO on the object detection/segmentation task on MS COCO dataset~\cite{lin2014microsoft}. 
We adopt the basic Mask R-CNN~\cite{he2017mask} and plug the LIO behind the Region Proposal Network, such that the structural information of each object can be well modeled. The SCL module can directly act on the object features after ROI pooling, thus the OEL module is disabled. We implemented the novel detection/segmentation network based on \textit{mmdetection}~\cite{chen2019mmdetection} toolbox and keep all hyper-parameters as default.

We apply the LIO module on the basic baseline of ResNet-50-C4 and a higher baseline of ResNeXt-101-FPN. The models are trained on COCO {\tt train2017} set and evaluated in the COCO {\tt val2017} set. We report the standard COCO metrics including $AP$, $AP_{50}$, $AP_{75}$ (averaged precision over multiple IoU thresholds), and $AP_S$, $AP_M$, $AP_L$ (AP across scales). Experimental results described in Table~\ref{tab:exp-detection} show that modeling structural compositions benefit object understanding and lead to better results on semantic segmentation. 
This demonstrated the effectiveness and generalization ability of our LIO for object structural compositions learning. Some examples of results by our basic ResNeXt-101-FPN and our approach are given in Fig.~\ref{fig:maskrcnn_visualization}.

\begin{table}[!ht]
\small
\begin{center}
    \begin{tabular}{|l|cc|}
    \hline
    \multicolumn{1}{|c|}{\multirow{2}{*}{Method}}  & \multicolumn{2}{c|}{Accuracy (\%)}  \\ \cline{2-3} 
    \multicolumn{1}{|c|}{} & CUB & CAR \\ \hline\hline
    ResNet-50~\cite{resnet} &  85.50  &  92.73      \\
    SCL          &  86.74  &  93.82 \\
    OEL         &  86.99  & 93.83 \\
    LIO          &  \textbf{87.31}  &  \textbf{93.89} \\ \hline\hline
    LIO w/ GM   &   87.37 & - \\
    \hline
\end{tabular}
\end{center}
    \caption{Ablation studies conducted on the proposed framework. ResNet-50: Basic ResNet-50 neural network trained by $\mathcal{L}_{cls}$. OEL: Model trained by $\mathcal{L}_{cls} + \alpha\mathcal{L}_{oel}$. SCL: Model trained by $\mathcal{L}_{cls} + \beta\mathcal{L}_{scl}$. LIO: Model trained by $\mathcal{L}$. GM: Ground truth semantic segmentation annotations.}
    \label{tab:exp-ablation}
\end{table}
\begin{figure}[!t]
    \centering
    \includegraphics[width=\linewidth,page=1]{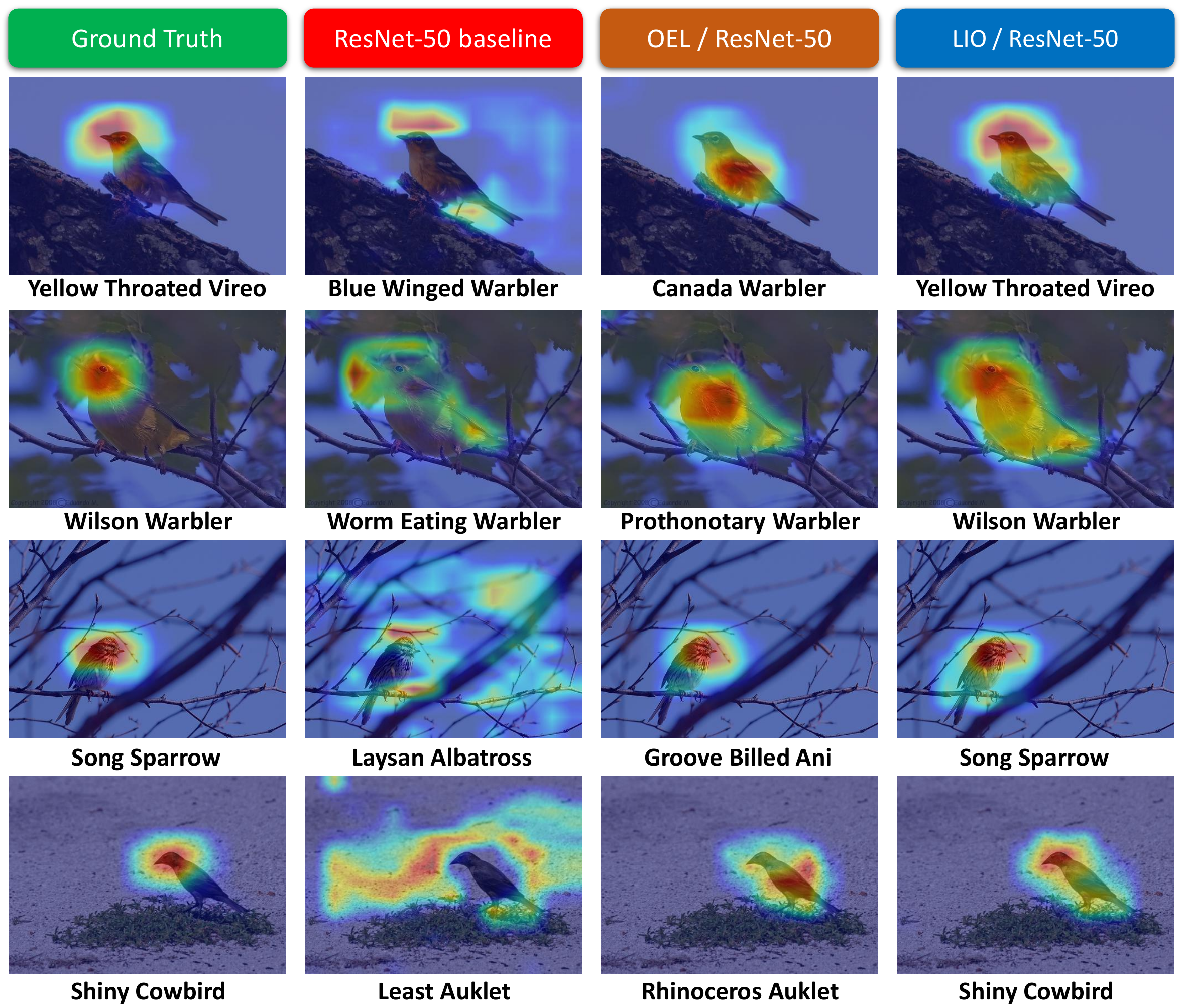}
    \caption{Visualization of feature maps by using OEL and SCL respectively. OEL enforce the backbone focus on object extent. SCL is helpful for not only searching discriminative region in object extent, but also completing the object extent localized by OEL.} 
    \label{fig:scl_effect}
\end{figure}

\subsection{Ablation Studies}

To demonstrate the effects of the OEL module and SCL module, we perform the module separation experiments on CUB~\cite{CUB} and CAR~\cite{CAR}. 
Both OEL and SCL act on the last stage feature map from the ResNet-50 backbone.
The results are shown in Table~\ref{tab:exp-ablation}. We can find that both modules improve performance significantly. In detail, as we show in Fig.~\ref{fig:scl_effect}, the SCL provides a principled way to learn the spatial structure, which is helpful for mining discriminative regions in an object. Moreover, the object extent can be localized by the OEL module according to the in-class region correlations and further defeats the negative influences from the diverse poses, appearance and background clutter. Together, the overall performance can be further improved owing to the complementary of nature SCL and OEL. 

Moreover, we also try to replace the pseudo semantic mask $M(I,\bm{I}')$ with the ground-truth mask for LIO. The results show that our learning based method can construct a high-quality semantic mask, which is even very close to the ground-truth mask (87.3\% vs. 87.4\% accuracy on CUB). 

\subsection{Discussions}
\begin{figure}[!t]
    \centerline{
    \includegraphics[width=\linewidth, page=1]{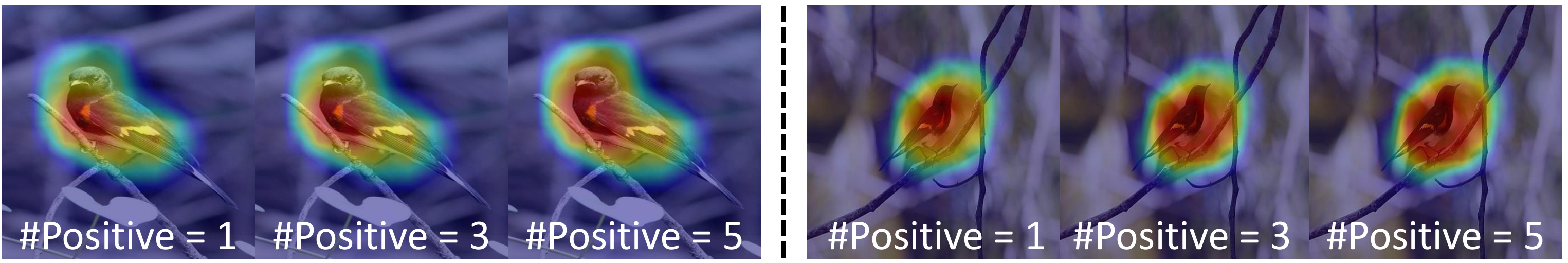}
    }
    \caption{Visualization of the changes
of pseudo segmentation masks given different number of positive images.}
\label{fig:positive_images} 
\end{figure}

\paragraph{\textbf{Number of Positive Images:}}
The number $P$ of positive images in a batch is an important parameter for the object-extent learning Module. We visualized the pseudo mask $M(I,\bm{I}')$ by given different number of positive images $P$ in Fig.~\ref{fig:positive_images}. We also evaluate our method on CUB and CAR with different numbers of positive images, and the recognition accuracy is shown in Table~\ref{tab:exp-num-positive}. 
With more positive images used, the framework gets better in structural learning and result in better performance. Finally, the performance will stop rising or falling and become steady. For a rigid object structure, such as CAR, we only need a few positive images for generating a reasonable pseudo extent mask. 

In general, feeding only one positive image may let the backbone learn fragmentary object extent for viewpoint diversity. The increase of $P$ leads to rapidly rising memory usage. Thus we use $P=3$ for experiments in this paper to trade-off between final performance and computation cost.
\begin{table}[!t]
\small
\begin{center}
    \begin{tabular}{|l|ccc|}
    \hline
    \multicolumn{1}{|c|}{\multirow{2}{*}{Dataset}}  & \multicolumn{3}{c|}{\# Positive Images}  \\ \cline{2-4} 
    \multicolumn{1}{|c|}{} & 1 & 3 & 5 \\ \hline\hline
    CUB & 86.83 & 87.31 & 87.30  \\
    CAR & 93.81 & 93.89 & 93.89  \\
    \hline
\end{tabular}
\end{center}
    \caption{The effect of the number of positive images on accuracy.}
    \label{tab:exp-num-positive}
\end{table}

\paragraph{\textbf{Model Efficiency:}}
During training time, our LIO introduced three additional layers besides the backbone network, including one convolutional layer in the OEL module, one convolutional layer and one fully-connected layer in the SCL module. 
For LIO/ResNet-50 (28x28), there are only 0.26  million new parameters introduced in our LIO, which is 1.01\% of \#Params of original ResNet-50. 

An important property is that both OEL and SCL modules can be disabled during testing. It means that the final classification model size is the same as the original backbone network. The baseline backbone network can be significantly improved without any computation overhead at inference time.

\section{Conclusions} \label{sec:conclusion}
In this paper, we proposed a Look-into-Object (LIO) framework to learn structure information for enhancing object recognition. We show that supervised object recognition could largely benefit from ``additional but free'' self-supervision, where geometric spatial relationship significantly rectifies the localization of discriminative regions and even result in better object detection and segmentation. Structural information, which was overlooked in prior literature, reliably prevents the network from falling into local confusion. Moreover, our plug-in style design can be widely adopted for injecting extra supervision into the backbone network without additional computational overhead for model deployment.
{\small
\bibliographystyle{ieee_fullname}
\bibliography{egbib}
}

\end{document}